\documentclass{article}
\usepackage{spconf,amsmath,graphicx}
\usepackage{booktabs}
\usepackage{multirow}
\usepackage{amssymb}
\usepackage{pifont}
\usepackage{tikz}
\usepackage{cite}
\usepackage{hyperref}
\hypersetup{colorlinks = true}
\usepackage[capitalise]{cleveref}
\usepackage{comment}
\usepackage{algorithmic}
\usepackage{xcolor}
\usepackage{url}
\usepackage{fancyhdr}

\fancypagestyle{firstpage}{
  \fancyhf{}                  

  \fancyfoot[C]{%
  \scriptsize
    © 2025 IEEE. Published in 2025 IEEE International Conference on Image Processing (ICIP), scheduled for 14-17 September 2025 in Anchorage, Alaska, USA. Personal use of this material is permitted. However, permission to reprint/republish this material for advertising or promotional purposes or for creating new collective works for resale or redistribution to servers or lists, or to reuse any copyrighted component of this work in other works, must be obtained from the IEEE. Contact: Manager, Copyrights and Permissions / IEEE Service Center / 445 Hoes Lane / P.O. Box 1331 / Piscataway, NJ 08855-1331, USA. Telephone: + Intl. 908-562-3966.%
  }%
}

\title{Towards Controllable Real Image Denoising with Camera Parameters}
\name{Youngjin Oh \qquad Junhyeong Kwon \qquad Keuntek Lee \qquad Nam Ik Cho}
\address{Department of ECE, INMC, Seoul National University, Seoul, Korea}
\begin{document}
%
\thispagestyle{firstpage}

\maketitle

\begin{abstract}
Recent deep learning-based image denoising methods have shown impressive performance; however, many lack the flexibility to adjust the denoising strength based on the noise levels, camera settings, and user preferences. In this paper, we introduce a new controllable denoising framework that adaptively removes noise from images by utilizing information from camera parameters. Specifically, we focus on ISO, shutter speed, and F-number, which are closely related to noise levels. We convert these selected parameters into a vector to control and enhance the performance of the denoising network. Experimental results show that our method seamlessly adds controllability to standard denoising neural networks and improves their performance. Code is available at \url{https://github.com/OBAKSA/CPADNet}.
\end{abstract}

\begin{keywords}
Adaptive layer normalization, camera parameters, controllable denoising, real image denoising.
\end{keywords}

\section{Introduction}
\label{sec:intro}

Image denoising is an essential process in photography that aims to restore a clear image from a noisy one. The primary source of noise in optical camera images is the sensor, which inevitably produces highly noisy images when sensor size and light intensity are reduced, as is common in many smartphone cameras~\cite{abdelhamed2018high}. Even high-end DSLR cameras, which are designed with larger sensors and wider apertures compared to smartphone cameras, still generate non-negligible noise. 

In recent years, there has been a rise in the development of deep learning-based image denoising methods, and several datasets~\cite{abdelhamed2018high,chen2018learning} were created through extensive and meticulous processes to support practical denoising research.
Additionally, due to the variability of noise levels in images for various reasons, numerous methods~\cite{zhang2018ffdnet,guo2019toward,soh2021deep,soh2022variational} have been developed to adjust the denoising strength flexibly to the noise level.

Among the methods, Guo \textit{et al.}~\cite{guo2019toward} proposed to estimate the noise level of a noisy image using a neural network to later use as an additional input of the denoising network.
Likewise, Soh \textit{et al.}~\cite{soh2021deep} splits the complex denoising problem into two relatively simple problems, each of which is solved by a sub-network. One sub-network is designed to estimate the noise level of a noisy image, while the other adaptively restores the clean image using both the estimated noise level and the noisy image itself. This approach also allows users to manually adjust the estimated noise level, enabling them to control the denoising strength according to their preferences.
These methods require synthetic data in training, as ground-truth noise levels are not available for real images.

In this work, we note that the noise level of real images is linked to several camera parameters, and thus we may provide more accurate estimates of the noise level by including them as inputs. We focus on three camera parameters available in the metadata of images: ISO, shutter speed (the reciprocal of exposure time), and F-number (aperture size). These values serve as valuable ground-truth information for real noise level, which we utilize for both training and inference.

Meanwhile, it has been noted that many neural networks using layer normalization show strong performance in image restoration, including real image denoising. Specifically, Chen \textit{et al.}~\cite{chen2022simple} have demonstrated that it is crucial to the performance and proposed a robust baseline architecture for image restoration. This finding emphasizes the need for further exploration of the potential of layer normalization.

In this context, we propose a simple yet effective method that enables users to control image denoising networks trained on real denoising data by using camera parameters as a prior. Specifically, we convert the key parameters related to real noise level---ISO, shutter speed, and F-number---into a vector, which is then integrated into the denoising network as a condition using adaptive layer normalization~\cite{peebles2023scalable}. Extensive experiments demonstrate that our method improves the performance of state-of-the-art denoising networks while offering controllability during the inference phase.

\section{Proposed Method}
\label{sec:proposed}
\subsection{Camera Parameters Related to Noise Level}
\label{sec:parameters}

\begin{figure*}
\centering
   \includegraphics[width=16.5cm]{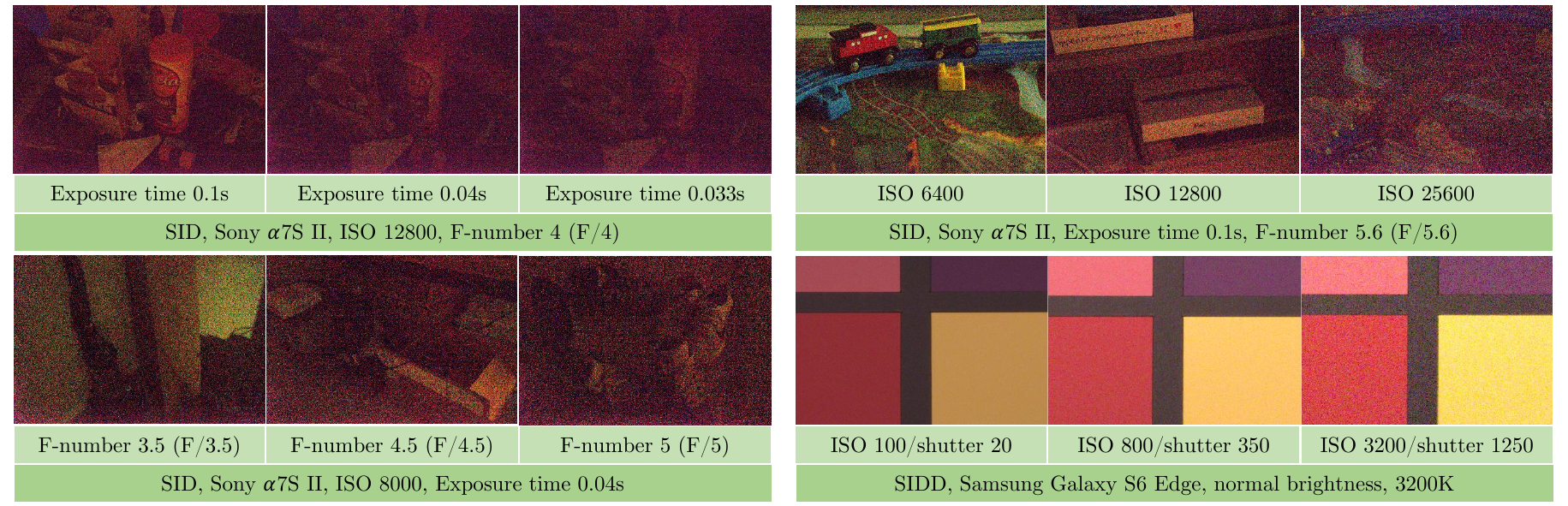}
   \hfil
\caption{Examples from SID~\cite{chen2018learning} and SIDD~\cite{abdelhamed2018high} that have been taken with the same camera settings except for one parameter. These examples demonstrate the impact of ISO, shutter speed, and F-number on noise intensity. It is evident that the noise intensity relative to the clean signal increases as the ISO increases, the shutter speed becomes faster, or the F-number increases.
}
\label{figure:iso}
\end{figure*}

Noise from optical sensors can generally be characterized using a heteroscedastic Gaussian model~\cite{foi2008practical,brooks2019unprocessing,conde2024toward}. When we have the intensity of a noise-free clean raw image represented as $x$, the noisy raw image $y$ is modeled as:
\begin{equation}
\label{eq1}
y \sim{\mathcal{N}(\mu=x, \sigma^2=\lambda_{read}+\lambda_{shot}x)},
\end{equation}
where the parameter $\lambda_{read}$ represents the signal-independent read noise, while the parameter $\lambda_{shot}$ represents the signal-dependent shot noise.

Our goal is to train a denoising network that effectively removes noise by utilizing specific camera parameters. Using \cref{eq1}, we select the parameters that are most closely related to the noise level of the captured image $y$, specifically ISO, shutter speed, and F-number. These three parameters are typically available in the EXIF metadata of images, which makes our approach practical. The impact of these camera parameters on image noisiness is illustrated in \cref{figure:iso}.

\noindent
\textbf{ISO.} 
The ISO setting is a key factor in determining the amount of noise in an image. It affects the sensor gains, which are linked to the noise parameters $\lambda_{read}$ and $\lambda_{shot}$ \cite{brooks2019unprocessing,wang2020practical}. When the ISO is increased, the image sensor output is amplified, enabling a faster shutter speed or a larger F-number at the same exposure when capturing a photograph. However, this amplification also increases noise levels in the image, as both $\lambda_{read}$ and $\lambda_{shot}$ are intensified.

\noindent
\textbf{Shutter Speed \& F-number.} 
Shutter speed and F-number also significantly influence the signal-to-noise ratio (SNR) of a captured image by directly affecting the amount of light that reaches the sensor.

Assuming the ISO is fixed, a camera with a slow shutter speed (long exposure time) or low F-number (wide aperture) will capture more light. In this case, the signal-independent read noise related to $\lambda_{read}$ remains constant, while the signal-dependent shot noise related to $\lambda_{shot}$ will increase as the light intensity of the clean image $x$ increases. However, shot noise increases at a slower rate than the clean signal $x$ itself, which improves the SNR and results in a cleaner image. Yet, long exposures can cause the sensor to heat up and induce thermal noise. On the other hand, a camera with a fast shutter speed (short exposure time) or high F-number (narrow aperture) will gather less light, reducing the signal's strength. Therefore, it decreases the SNR, making the noise more noticeable.

\begin{figure*}[t]
\centering
   \includegraphics[width=16.0cm]{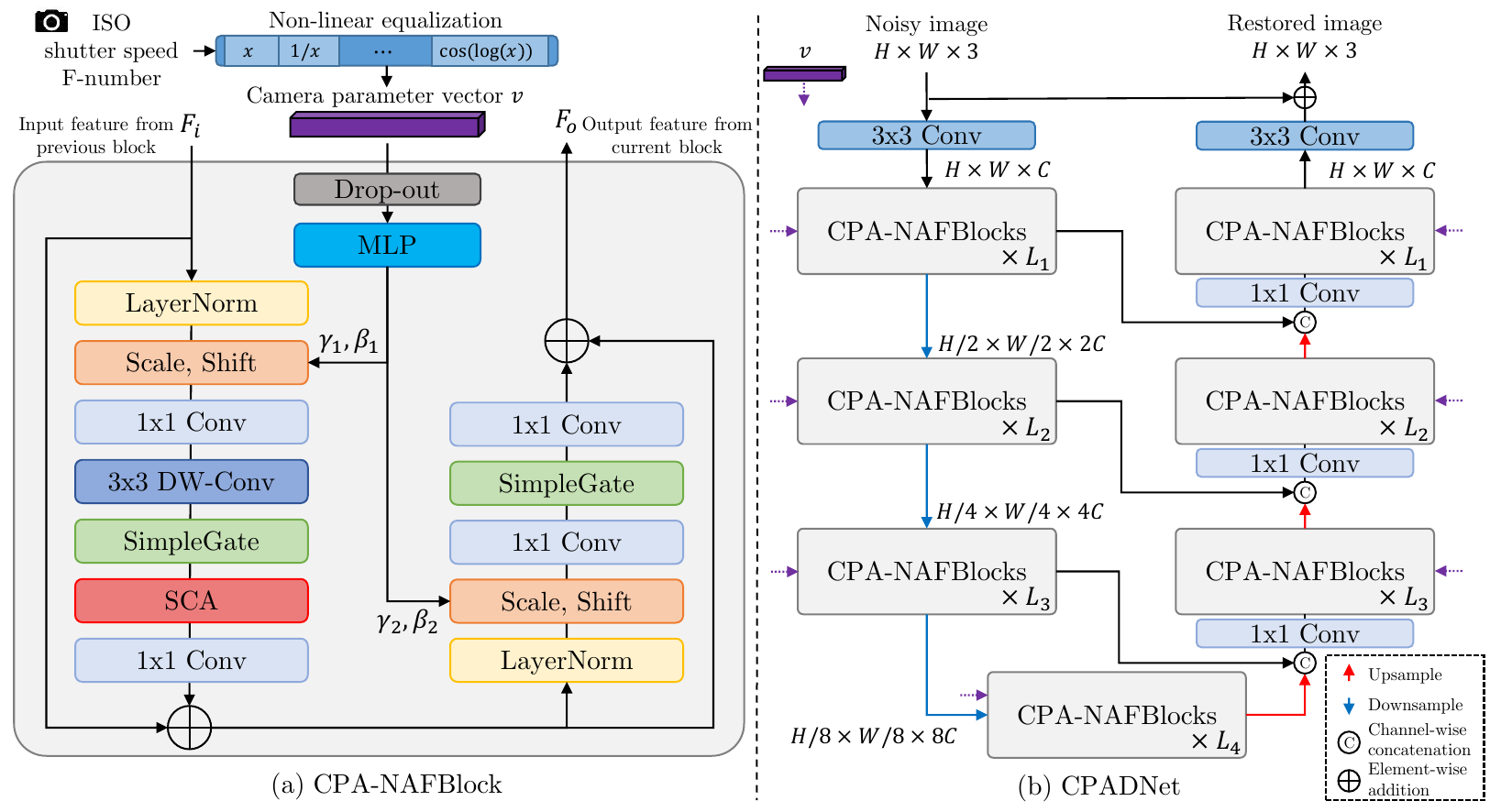}
   \hfil
\caption{(a) Proposed CPA-NAFBlock, which is modified from NAFBlock~\cite{chen2022simple} with adaptive layer normalization~\cite{peebles2023scalable}, and (b) CPADNet, a denoising network composed of CPA-NAFBlocks. CPADNet is designed as a U-shaped network with skip-connections and a hierarchical encoder-decoder architecture~\cite{wang2022uformer} for its effectiveness in image restoration.}
\label{figure:adaln}
\end{figure*}

\noindent
\textbf{Exposure Algorithm.} 
In practice, ISO and shutter speed/F-number are adjusted automatically by an exposure algorithm according to the scene illumination, mostly in \textit{aperture-priority} or \textit{shutter-priority} mode~\cite{abuolaim2020defocus}. 
Unlike DSLR cameras which have adjustable apertures, most smartphone cameras have a fixed aperture size, and consequently, the exposure algorithm operates with only ISO and shutter speed.

In a typical situation, ISO and shutter speed/F-number are usually positively correlated. For instance, using a high ISO amplifies the light intensity, allowing for a fast shutter speed or a high F-number. On the other hand, a low ISO could cause the sensor to amplify the signal insufficiently, requiring a slower shutter speed or low F-number.

However, an image captured with a slow shutter speed is more likely to be affected by motion blur, while a low F-number indicates a wide aperture and a shallow depth of field (DoF), causing defocus deblur~\cite{abuolaim2020defocus}. Therefore, careful consideration is needed when balancing the camera parameters, including their impact on noise and exposure; otherwise, other types of unwanted degradation could be induced.



\subsection{Camera Parameter-Adaptive Denoising Network}
Our method focuses on conditioning a denoising network with noise-related camera parameters and controlling its denoising strength. By providing the network with accurate camera parameters during training, it can effectively learn how to remove noise from real images with varying noise levels. For instance, during training, providing the ISO alongside a noisy image taken with a high ISO will guide the network to remove noise more aggressively compared to an image taken with a low ISO. Similarly, an image captured with a slow shutter speed or low F-number will indicate less noise for the network to remove than an image captured with a fast shutter speed or high F-number.

However, there are some challenges in implementing our ideas; camera parameters vary in non-uniform steps~\cite{kim2024paramisp}, and the impact of the parameters on noise level or denoising strength is not simply linear. Consequently, in order to model the non-linear impact, we follow \cite{kim2024paramisp} and apply non-linear equalization to the camera parameters and choose $x$, $1/x$, $\sqrt{x}$, $x^{-1/2}$, $x^{1/4}$, $x^{-1/4}$, $\mathrm{log}(x)$, $\mathrm{sin}(\mathrm{log}(x))$, and $\mathrm{cos}(\mathrm{log}(x))$ as the non-linear functions. We apply each functions to the camera parameters and form 9-dimensional vectors, and the mapped results are normalized to be in the range of $[0, 1]$.
As we utilize three camera parameters, each individual vectors are concatenated to form a camera parameter vector $v\in\mathbb{R}^{27}$, which briefly represents the noise level of an image.

To apply our method for smartphone data that do not provide F-numbers, we alternatively one-hot encode the smartphone types and produce a vector with a single embedding layer to represent the different noise of the various sensors instead of the non-linearly mapped F-number vector.

In order to exploit this camera parameter vector which contains prior knowledge of the noise level as a condition at the training phase and also to harness the denoising network with controllability at inference, our method adopts adaptive layer normalization (adaLN)~\cite{peebles2023scalable}. As discussed in \cref{sec:intro}, recent image restoration networks have found layer normalization beneficial due to its stable training and superior performance compared to other normalization methods~\cite{chen2022simple}. By incorporating the camera parameter vector $v$ into adaLN layers, we can condition each block of the network with channel-wise affine parameters $\beta$ and $\gamma$. This design allows the network to adaptively modify its denoising strength based not only the noisy image itself but also on the camera parameters.

\begin{figure}[t]
\centering
   \includegraphics[width=7.0cm]{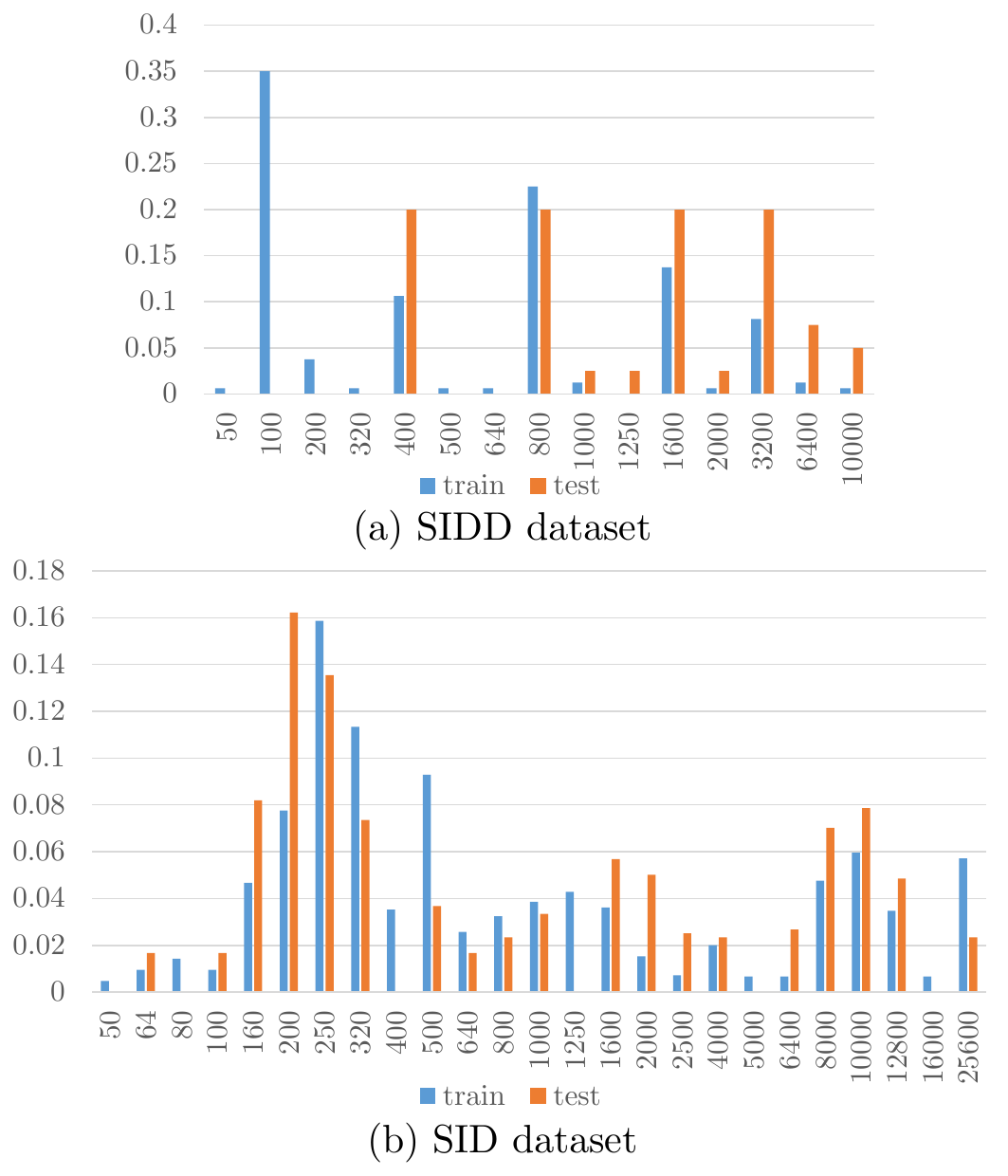}
   \hfil
\caption{Distribution of ISO value for SID and SIDD datasets.}
\label{figure:distribution}
\end{figure}

\begin{figure*}[t]
\centering
   \includegraphics[width=17.0cm]{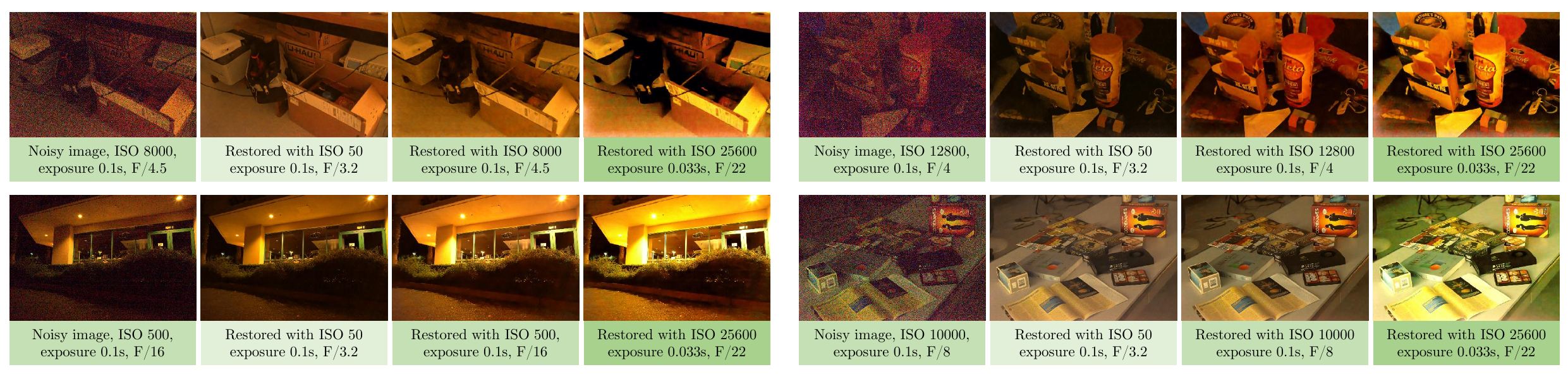}
   \hfil
\caption{Results of controllable denoising with CPADNet on full-frame SID test set.}
\label{figure:qualitative_sid}
\end{figure*}

\begin{figure*}[t]
\centering
   \includegraphics[width=17.0cm]{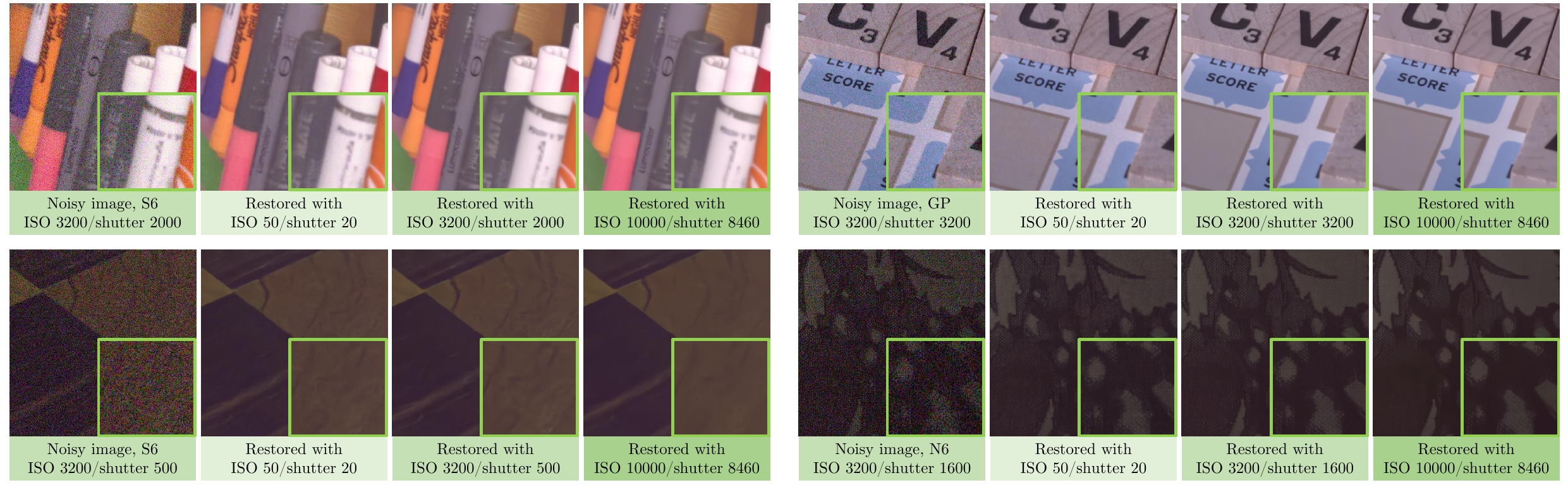}
   \hfil
\caption{Results of controllable denoising with CPADNet on full-frame SIDD test set. Examples are cropped for closer analysis.}
\label{figure:qualitative_sidd}
\end{figure*}

\begin{table*}[t]
\centering
\caption{Evaluation of our proposed method applied to various restoration networks with SID and SIDD datasets.}
\resizebox{0.9\textwidth}{!}{
\begin{tabular}{c|cc|cc|cc|cc|cc}
\hline
\multirow{2}{*}{Method} & \multirow{2}{*}{Params} & \multirow{2}{*}{MACs} & \multicolumn{2}{c|}{SID Sony (patch)} & \multicolumn{2}{c|}{SID Sony (full)} & \multicolumn{2}{c|}{SIDD valid (patch)} & \multicolumn{2}{c}{SIDD test (patch)} \\ \cline{4-11} 
                        &                         &                       & PSNR              & SSIM              & PSNR             & SSIM              & PSNR               & SSIM               & PSNR              & SSIM              \\ \hline
Baseline                & 6.67M                   & 18.53G                & 24.58             & 0.8635            & 23.62            & 0.8633            & 39.41              & 0.9569             & 39.57             & 0.9120            \\
CPADNet                 & 9.45M                   & 18.53G                & 25.57             & 0.8729            & 24.06            & 0.8652            & 39.43              & 0.9574             & 39.59             & 0.9131            \\ \hline
Restormer               & 11.72M                  & 64.29G                & 25.01             & 0.8579            & -                & -                 & 39.60              & 0.9586             & 39.77             & 0.9149            \\
CPA-Restormer           & 15.06M                  & 64.30G                & 25.81             & 0.8758            & -                & -                 & 39.62              & 0.9589             & 39.79             & 0.9151            \\ \hline
\end{tabular}
\label{table:evaluation}
}
\end{table*}

To demonstrate that our approach introduces controllability and enhances performance of existing denoising networks, we first select a robust denoising network as our baseline. Specifically, we construct our denoising baseline network using NAFBlocks~\cite{chen2022simple}, known for their simple yet outstanding performance in various image restoration tasks.
\cref{figure:adaln} depicts this structure, named Camera Parameter-Adaptive Denoising Network (CPADNet). We replace the standard layer normalization in the original NAFBlocks with adaptive layer normalization to provide conditional information about the noise level for the network to utilize. The MLP layer consists of two linear layers with SiLU~\cite{elfwing2018sigmoid} activation function in between. Dropout with a probability of 0.2 is introduced before calculating the affine parameters to prevent overfitting to certain parameter combinations.
Our method is simple to implement and can be easily integrated as a plug-and-play module in other restoration networks. Besides CPADNet, we also present results of applying our method to a Transformer-based network~\cite{zamir2022restormer} to support this claim.

\section{Experimental Results}
\label{sec:experiments}
\subsection{Dataset}
\label{sec:dset}
We train and evaluate our method using the Sony subset of the SID~\cite{chen2018learning} and SIDD-medium datasets~\cite{abdelhamed2018high}. 

\noindent
\textbf{SID.} 
The Sony subset of SID dataset is taken with Sony $\alpha$7S II camera, and from the total of 2697 short-/long-exposure RAW image pairs, we use 2099 pairs as training data and 598 pairs as test data~\cite{cai2023retinexformer}. The dataset is used in both low-light image enhancement~\cite{cai2023retinexformer} and denoising tasks~\cite{jin2024mipi}. We obtain noisy/clean RGB images by applying Rawpy's postprocess() function to short-/long-exposure RAW image pairs. The camera parameters ISO, shutter speed, and F-number for each noisy image are provided as text files.

\noindent
\textbf{SIDD.} The SIDD dataset consists of a total of 400 image pairs, divided into 320 for training, 40 for validation, and 40 for testing. The ground truth of the test set is undisclosed to the public. The dataset contains various noisy images taken with 5 different smartphone cameras. Each noisy image is accompanied by labels indicating the camera and environmental settings, such as ISO, shutter speed, scene brightness, illuminant temperature, and camera type. We extract these labels from the official dataset webpage.

\subsection{Implementation Details}
\label{sec:details}
We implement the networks using the official code released by the authors~\cite{chen2022simple} and train CPADNet with and without our method on an NVIDIA RTX 3090 GPU. 
For the SID and SIDD training set, we crop overlapping patches of size $256\times 256$ with stride of 196 and 180, respectively. We randomly create a mini-batch of size 8 for each iteration. The networks are trained in an end-to-end manner with $L_1$ loss using Adam optimizer with initial learning rate of $2\times10^{-4}$, gradually decaying to $1\times10^{-6}$ with cosine annealing scheduling for 200K iterations. The number of channels $C$ of the first layer of the network is $32$, and the number of blocks  $L_1$, $L_2$, $L_3$, and $L_4$ in each level of the network are $[4,4,4,8]$, respectively.

\subsection{Evaluation of Our Method}
The denoising performance is measured using PSNR and SSIM as the metrics. Additionally, we calculate the MACs by processing a $256\times256$-sized input and measure the number of parameters to assess the computational efficiency. We also measure the results for SID using non-overlapping patches of full-frame test images for methods~\cite{zamir2022restormer} that cannot handle high-resolution input. For SIDD~\cite{abdelhamed2018high}, we conduct a quantitative evaluation using the validation set of the dataset, and the test set uploaded to Kaggle. For the qualitative evaluation, we use the full-frame images for both SID and SIDD as this is the practical restoration setting.

\Cref{table:evaluation} shows the result of the experiments. Our proposed method, denoted as CPADNet, incorporates the noise-related camera parameters as additional input to the baseline. CPADNet demonstrates improved performance in both PSNR and SSIM metrics, highlighting the effectiveness of our method. 
We have also applied our method on Restormer~\cite{zamir2022restormer} with initial width 32, referred to as CPA-Restormer, using a mini-batch size of 2 to accommodate training memory limitations. The outcomes demonstrate that our method can be effectively utilized with another state-of-the-art network. Our method does increase parameters but is computationally efficient, as we can implement it with minimal increases in MACs.

Our method improves the denoising performance more profoundly for the SID dataset than the SIDD dataset. We believe this is due to the discrepancy between the noise level distribution in the training and test set of the datasets. \cref{figure:distribution} shows that the distribution of ISO values in the training and test set of SID is more consistent when compared to SIDD. Noisy images with rare ISO values were excluded as outliers in works on a related topic of noise generation, for instance, \cite{maleky2022noise2noiseflow}, when using SIDD. However, our method includes all the ISO values and demonstrates that it is robust and enhances the performance even with outlier noise levels.

\subsection{Controllability and Dependency on Datasets}
\cref{figure:qualitative_sidd,figure:qualitative_sid} demonstrates the controllability of our method, and the camera parameters at inference have been chosen to highlight the weakest and strongest denoising strength. As mentioned earlier, our network is designed to reduce significant noise in images captured with high noise-related camera parameters and vice versa. This implies that the network will reduce less noise if given a lower ISO, shutter speed, or F-number than the original camera settings provided by the metadata. Conversely, in the opposite situation, the network will remove more noise.

Following most supervised deep learning methods, our method is also influenced by the noise characteristics of the training data, which are SID and SIDD in our work. In both cases, users have the ability to control the denoiser and generate a restored output according to their preference during the inference process with our method.

\noindent
\textbf{SID.}
A large percentage of the noisy images in SID are degraded by heavy noise in low-illumination settings to the point that it alters the image's overall color to a magenta or purple shade. Influenced by this characteristic, our method is guided to control the overall color of the image along with the noise and texture, as displayed in \cref{figure:qualitative_sid}.

\noindent
\textbf{SIDD.}
SIDD consists of images of various noise levels, but even the most substantial noise is weak compared to the SID dataset. From this data composition, our method learns fine-grained controllability of the textures without changing the overall color. 
As shown in \cref{figure:qualitative_sidd}, reducing the denoising strength on a noisy image enhances the details but removes less noise. Contrarily, heavily applying the denoising process to a noisy image results in a smoother output with less texture. 

\section{Conclusion}
\label{sec:conclusion}
We have developed an effective method to enhance and control the performance of a denoising network by using noise-related camera parameters. To achieve this, we have created a vector containing information about the noise level and utilize adaptive layer normalization to empower the network with this vector. Experiments demonstrate that our method improves the performance of the denoising network by leveraging the parameters as prior information. Additionally, our proposed method can adjust the denoising strength by using the input camera parameters as a condition during the inference phase based on the user's preference.

\noindent
\textbf{Acknowledgment} 
This research was supported by Samsung Electronics Co., Ltd., and in part by the BK21 FOUR program of the Education and Research Program for Future ICT Pioneers, Seoul National University in 2025.


\begin{thebibliography}{10}

\bibitem{abdelhamed2018high}
Abdelrahman Abdelhamed, Stephen Lin, and Michael~S Brown,
\newblock ``A high-quality denoising dataset for smartphone cameras,''
\newblock in {\em Proceedings of the IEEE conference on computer vision and pattern recognition}, 2018, pp. 1692--1700.

\bibitem{chen2018learning}
Chen Chen, Qifeng Chen, Jia Xu, and Vladlen Koltun,
\newblock ``Learning to see in the dark,''
\newblock in {\em Proceedings of the IEEE conference on computer vision and pattern recognition}, 2018, pp. 3291--3300.

\bibitem{zhang2018ffdnet}
Kai Zhang, Wangmeng Zuo, and Lei Zhang,
\newblock ``Ffdnet: Toward a fast and flexible solution for cnn-based image denoising,''
\newblock {\em IEEE Transactions on Image Processing}, vol. 27, no. 9, pp. 4608--4622, 2018.

\bibitem{guo2019toward}
Shi Guo, Zifei Yan, Kai Zhang, Wangmeng Zuo, and Lei Zhang,
\newblock ``Toward convolutional blind denoising of real photographs,''
\newblock in {\em Proceedings of the IEEE/CVF conference on computer vision and pattern recognition}, 2019, pp. 1712--1722.

\bibitem{soh2021deep}
Jae~Woong Soh and Nam~Ik Cho,
\newblock ``Deep universal blind image denoising,''
\newblock in {\em 2020 25th International Conference on Pattern Recognition (ICPR)}. IEEE, 2021, pp. 747--754.

\bibitem{soh2022variational}
Jae~Woong Soh and Nam~Ik Cho,
\newblock ``Variational deep image restoration,''
\newblock {\em IEEE Transactions on Image Processing}, vol. 31, pp. 4363--4376, 2022.

\bibitem{chen2022simple}
Liangyu Chen, Xiaojie Chu, Xiangyu Zhang, and Jian Sun,
\newblock ``Simple baselines for image restoration,''
\newblock in {\em European Conference on Computer Vision}. Springer, 2022, pp. 17--33.

\bibitem{peebles2023scalable}
William Peebles and Saining Xie,
\newblock ``Scalable diffusion models with transformers,''
\newblock in {\em Proceedings of the IEEE/CVF International Conference on Computer Vision}, 2023, pp. 4195--4205.

\bibitem{foi2008practical}
Alessandro Foi, Mejdi Trimeche, Vladimir Katkovnik, and Karen Egiazarian,
\newblock ``Practical poissonian-gaussian noise modeling and fitting for single-image raw-data,''
\newblock {\em IEEE transactions on image processing}, vol. 17, no. 10, pp. 1737--1754, 2008.

\bibitem{brooks2019unprocessing}
Tim Brooks, Ben Mildenhall, Tianfan Xue, Jiawen Chen, Dillon Sharlet, and Jonathan~T Barron,
\newblock ``Unprocessing images for learned raw denoising,''
\newblock in {\em Proceedings of the IEEE/CVF Conference on Computer Vision and Pattern Recognition}, 2019, pp. 11036--11045.

\bibitem{conde2024toward}
Marcos~V Conde, Florin Vasluianu, and Radu Timofte,
\newblock ``Toward efficient deep blind raw image restoration,''
\newblock in {\em 2024 IEEE International Conference on Image Processing (ICIP)}. IEEE, 2024, pp. 1725--1731.

\bibitem{wang2020practical}
Yuzhi Wang, Haibin Huang, Qin Xu, Jiaming Liu, Yiqun Liu, and Jue Wang,
\newblock ``Practical deep raw image denoising on mobile devices,''
\newblock in {\em European Conference on Computer Vision}. Springer, 2020, pp. 1--16.

\bibitem{wang2022uformer}
Zhendong Wang, Xiaodong Cun, Jianmin Bao, Wengang Zhou, Jianzhuang Liu, and Houqiang Li,
\newblock ``Uformer: A general u-shaped transformer for image restoration,''
\newblock in {\em Proceedings of the IEEE/CVF conference on computer vision and pattern recognition}, 2022, pp. 17683--17693.

\bibitem{abuolaim2020defocus}
Abdullah Abuolaim and Michael~S Brown,
\newblock ``Defocus deblurring using dual-pixel data,''
\newblock in {\em Computer Vision--ECCV 2020: 16th European Conference, Glasgow, UK, August 23--28, 2020, Proceedings, Part X 16}. Springer, 2020, pp. 111--126.

\bibitem{kim2024paramisp}
Woohyeok Kim, Geonu Kim, Junyong Lee, Seungyong Lee, Seung-Hwan Baek, and Sunghyun Cho,
\newblock ``Paramisp: Learned forward and inverse isps using camera parameters,''
\newblock in {\em Proceedings of the IEEE/CVF Conference on Computer Vision and Pattern Recognition}, 2024, pp. 26067--26076.

\bibitem{elfwing2018sigmoid}
Stefan Elfwing, Eiji Uchibe, and Kenji Doya,
\newblock ``Sigmoid-weighted linear units for neural network function approximation in reinforcement learning,''
\newblock {\em Neural networks}, vol. 107, pp. 3--11, 2018.

\bibitem{zamir2022restormer}
Syed~Waqas Zamir, Aditya Arora, Salman Khan, Munawar Hayat, Fahad~Shahbaz Khan, and Ming-Hsuan Yang,
\newblock ``Restormer: Efficient transformer for high-resolution image restoration,''
\newblock in {\em Proceedings of the IEEE/CVF conference on computer vision and pattern recognition}, 2022, pp. 5728--5739.

\bibitem{cai2023retinexformer}
Yuanhao Cai, Hao Bian, Jing Lin, Haoqian Wang, Radu Timofte, and Yulun Zhang,
\newblock ``Retinexformer: One-stage retinex-based transformer for low-light image enhancement,''
\newblock in {\em Proceedings of the IEEE/CVF International Conference on Computer Vision}, 2023, pp. 12504--12513.

\bibitem{jin2024mipi}
Xin Jin, Chunle Guo, Xiaoming Li, Zongsheng Yue, Chongyi Li, Shangchen Zhou, Ruicheng Feng, Yuekun Dai, Peiqing Yang, Chen~Change Loy, et~al.,
\newblock ``Mipi 2024 challenge on few-shot raw image denoising: Methods and results,''
\newblock in {\em Proceedings of the IEEE/CVF Conference on Computer Vision and Pattern Recognition}, 2024, pp. 1153--1161.

\bibitem{maleky2022noise2noiseflow}
Ali Maleky, Shayan Kousha, Michael~S Brown, and Marcus~A Brubaker,
\newblock ``Noise2noiseflow: Realistic camera noise modeling without clean images,''
\newblock in {\em Proceedings of the IEEE/CVF Conference on Computer Vision and Pattern Recognition}, 2022, pp. 17632--17641.

\end{thebibliography}
\end{document}